\documentclass[11pt]{article}

\usepackage[preprint]{acl}

\usepackage{times}
\usepackage{latexsym}

\usepackage[T1]{fontenc}

\usepackage[utf8]{inputenc}

\usepackage{microtype}

\usepackage{inconsolata}

\usepackage{graphicx}
\usepackage{tabularx}
\usepackage{booktabs}
\usepackage{amssymb}
\usepackage{float}
\usepackage{stfloats}
\usepackage{chngcntr} 
\newcommand\blfootnote[1]{%
  \begingroup
  \renewcommand\thefootnote{}\footnote{#1}%
  \addtocounter{footnote}{-1}%
  \endgroup
}

\title{\name: Open Source AutoML Library for NLU}

\author{
  \textbf{Grigory Arshinov\textsuperscript{1}},
  \textbf{Aleksandr Boriskin\textsuperscript{*,1,2}},
  \textbf{Sergey Senichev\textsuperscript{*,1,2}},
\\
  \textbf{Ayaz Zaripov\textsuperscript{1}},
  \textbf{Daria Galimzianova\textsuperscript{1,3}},
  \textbf{Daniil Karpov\textsuperscript{1,$\dagger$}},
  \textbf{Leonid Sanochkin\textsuperscript{1,$\dagger$}}
\\
\\
  \textsuperscript{1}MWS AI,
  \textsuperscript{2}ITMO University,
  \textsuperscript{3}MBZUAI
\\
  \small{
    \textbf{Correspondence:} \href{mailto:g.arshinov@mts.ai}{\texttt{g.arshinov@mts.ai}}
  }
}

\newcommand{\name}{OpenAutoNLU}
 
\begin{document}
\maketitle
\blfootnote{* Equal contribution.}
\blfootnote{$\dagger$ Work done while at MWS AI}

\begin{abstract}
\name{} is an open-source automated machine learning library for natural 
language understanding (NLU) tasks, covering both text classification and named entity recognition (NER). Unlike existing solutions, we introduce 
data-aware training regime selection that requires no manual configuration from 
the user. The library also provides integrated data quality diagnostics, 
configurable out-of-distribution (OOD) detection, and large language model (LLM) features, all within a minimal low-code API. \name{} source code is available here \footnote{https://github.com/mts-ai/OpenAutoNLU}, the demo app is accessible here \href{https://openautonlu.dev}{https://openautonlu.dev}.

\end{abstract}

\section{Introduction}

Text classification and NER are foundational tasks in natural language processing (NLP), underpinning applications from intent detection and sentiment analysis to information extraction and document categorization. Despite their ubiquity, deploying effective models for these tasks remains challenging: practitioners must navigate a complex landscape of competing approaches—full fine-tuning of pretrained transformers, few-shot learning methods, classical machine learning with embeddings—each suited to different data regimes and resource constraints. Further complications arise from data quality issues, hyperparameter sensitivity, and the need for OOD detection in production settings. Automated machine learning (AutoML) offers a promising solution by automating model selection, hyperparameter tuning, and pipeline design, yet existing frameworks exhibit significant gaps when applied to NLP workloads.

\begin{figure}[H]
    \centering
    \includegraphics[width=\columnwidth,height=0.72\textheight,keepaspectratio]{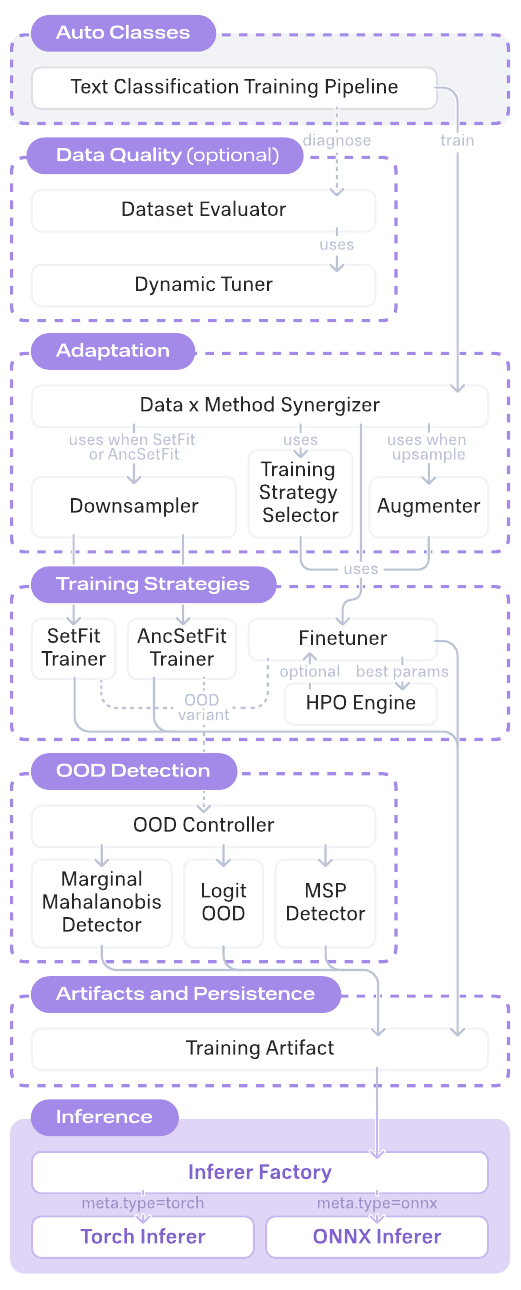}
    \caption{Text Classification Training Pipeline and Inference Pipeline flow}
    \label{fig:architecture}
\end{figure}


We identify two critical gaps in the current AutoML landscape for NLP. 
\textit{First}, ease of use: many frameworks require non-trivial configuration, expose complex abstractions (e.g., table-based predictors or modality-specific modules), and lack simple, unified interfaces for common NLP tasks. \textit{Second}, NLP-centric design: existing systems do not natively integrate (a) automatic selection of training regimes (full fine-tuning vs. few-shot methods) based on dataset size and label distribution, (b) text-specific data-quality assessment, and (c) unified support for both text classification and NER within a single coherent API. 


In this work, we present \name, an open-source AutoML library designed specifically for NLP. \name{} addresses the identified gaps through a text-first architecture that emphasizes simplicity, automation, and reproducibility, and demonstrates competitive or superior performance compared to existing AutoML frameworks in standard intent classification benchmarks. The library provides the following key \textbf{contributions}:
Automatic training-regime selection, Integrated data-quality tools, Configurable OOD detection, Unified API for classification and NER, Minimal configuration.

\section{Background}

\newcommand{\thead}[1]{\begin{tabular}[c]{@{}c@{}}#1\end{tabular}}

\begin{table*}[t]
\centering
\small
\setlength{\tabcolsep}{4pt}
\renewcommand{\arraystretch}{1.15}
\begin{tabularx}{\textwidth}{@{} l *{6}{>{\centering\arraybackslash}X} @{}}
\toprule
 & \textbf{H2O}
 & \thead{\textbf{Light}\\\textbf{AutoML}}
 & \thead{\textbf{Auto}\\\textbf{Gluon}}
 & \thead{\textbf{Auto}\\\textbf{Intent}}
 & \thead{\textbf{OpenAuto}\\\textbf{NLU}} \\
\midrule
Approach &
TAML \& Word2Vec &
TAML \& embeddings &
Encoder finetuning &
Embeddings &
Encoder finetuning \\
\addlinespace[3pt]
Scarce train data &
$\times$ &
Has small-data modes &
Adaptable &
Adapted for small datasets &
Designed for classes with $\geq$2 examples \\
\addlinespace[3pt]
Custom search pipeline &
Flexible API &
HP presets &
HP presets &
Presets \& customizable configs &
Universal customizable preset \\
\addlinespace[3pt]
LLM Integrations &
$\times$ &
$\times$ &
$\times$ &
Zero-shot classification &
Train data augmentations and test set generation\\
\addlinespace[3pt]
OOD detection &
$\times$ &
$\times$ &
$\times$ &
Train examples required, one method &
Train examples are optional, configurable \\
\addlinespace[3pt]
Optimized for NLP production &
Requires preprocessing; Java, Python, R support &
$\times$ &
\checkmark &
$\times$ &
ONNX native; inference in two lines \\
\addlinespace[3pt]
Data quality evaluation &
$\times$ &
$\times$ &
$\times$ &
$\times$ & 
\checkmark \\
\addlinespace[3pt]
Named Entity Recognition &
$\times$ &
$\times$ &
\checkmark &
$\times$ &
\checkmark \\
\addlinespace[3pt]
Embedding prompting &
$\times$ &
$\times$ &
$\times$ &
\checkmark &
\checkmark \\
\addlinespace[3pt]
\bottomrule
\end{tabularx}
\caption{Comparison of AutoML frameworks for NLP tasks.}
\label{tab:automl_comparison}
\end{table*}

\begin{figure}[h]
    \centering
    \includegraphics[width=1\linewidth]{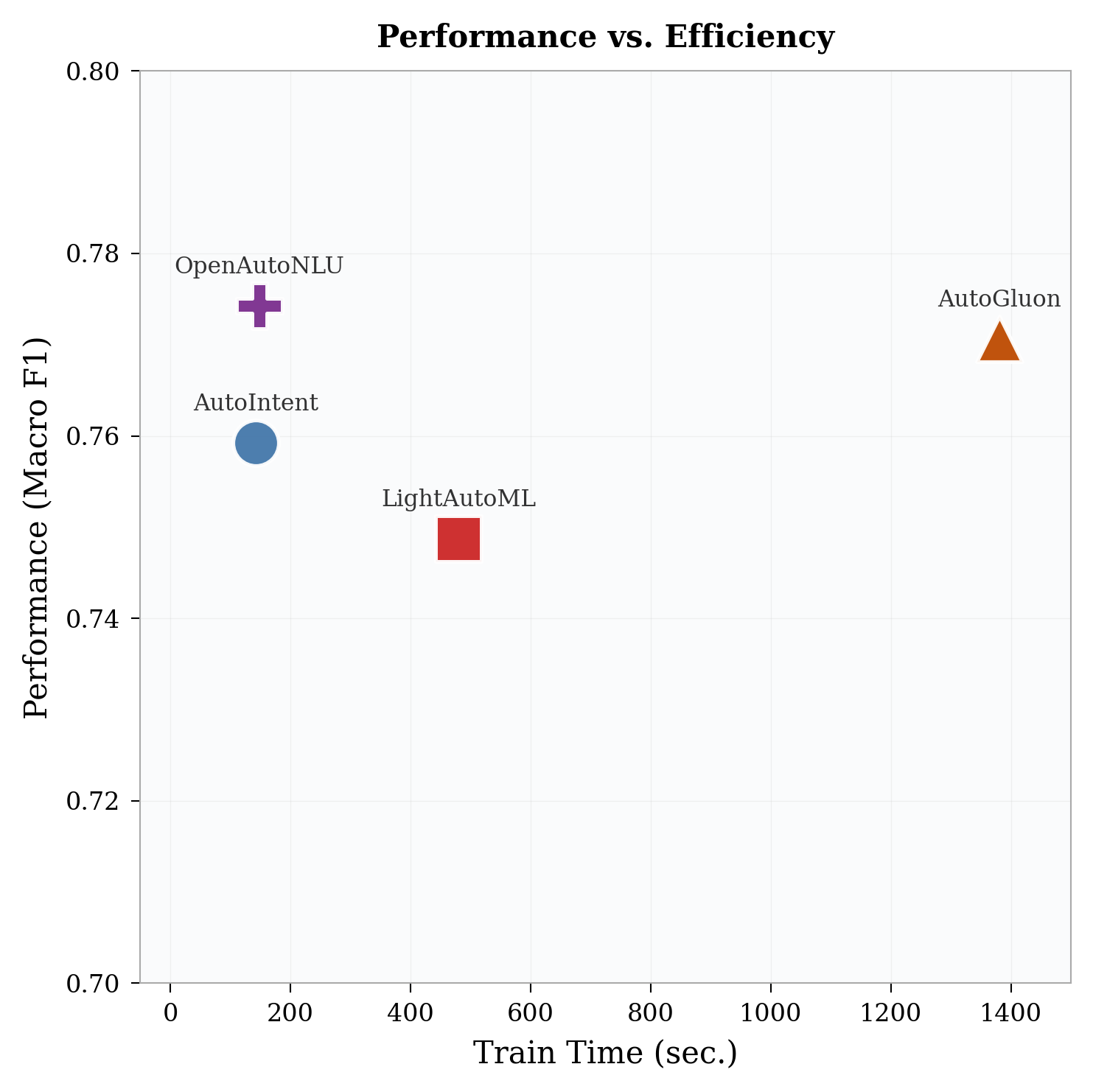}
    \caption{This graph illustrates the ratio between performance in macro F1 on classification tasks and time in seconds took to train the solution. Measures are averaged between four different text classification datasets.}
    \label{fig:perfomance_vs_efficiency}
\end{figure}

Recently, large language models (LLMs) have also been proposed as general-purpose solutions for text classification and intent understanding, often via prompting or lightweight adaptation. While such models can provide strong zero- and few-shot performance \cite{wang2024adaptablereliabletextclassification}, they typically incur substantial computational and monetary costs, as well as higher inference latency due to their size and deployment requirements (e.g., multiple GPU-backed serving or external API calls). These characteristics make LLM-based approaches less suitable for many practical scenarios where models must be deployed at scale \cite{vajjala2025textclassificationllmera}, integrated on-premise, or executed under strict latency and resource constraints. In contrast, the AutoML frameworks considered in this work aim to deliver competitive performance using comparatively lightweight architectures and training pipelines, which are more amenable to efficient, low-latency deployment.

To our knowledge, some solutions for AutoML have previously been introduced. For example, AutoIntent \cite{alekseev2025autointent} is an AutoML framework specifically designed for intent classification. It follows an embedding-centric design, where pretrained sentence encoders are combined with either classical classifiers 
or neural models, followed by automatic decision threshold optimization. AutoIntent supports \textbf{supervised} out-of-distribution (OOD) detection, multi-label classification, and multiple training presets that trade off quality and computational cost. In our experiments, we evaluated classic-light (default), classic-medium, nn-medium, nn-heavy and transformers-heavy AutoIntent presets.

 AutoGluon \cite{erickson2020autogluon} is a general-purpose AutoML framework originally developed for tabular data, relying heavily on multi-layer ensembling, stacking, and bagging of heterogeneous models (tree-based models, linear models, and neural networks). 

LightAutoML \cite{vakhrushev2021lightautoml} and H2O AutoML \cite{ledell2020h2o} are also mainly designed for tabular data. Text inputs are handled by simple Word2Vec-based vectorization and subsequently processed by standard tabular models. All experiments with these frameworks were used with default configurations.


Feature overview for all frameworks can be found in the Table~\ref{tab:automl_comparison}.

\section{\name}
\name{} is a library in the Python language that is designed to be used in low-code environments. Its simple API allows for quick NLU-model prototyping as well as for robust and flexible production model training. Architecturally, it is a collection of classes that are chained by auto pipelines. The diagram can be found in the Figure~\ref{fig:architecture}.

\textbf{Novelty}
The principal novelty of \name{} lies in its \emph{resolution of the data-aware method}: rather than exposing the user to a menu of algorithms and hyperparameter grids, the library inspects the label distribution of the supplied dataset and deterministically selects the training method that best fits the data regime.
Three regimes are distinguished by the minimum per-class sample count $n_{\min}$ (which were computed on our internal benchmarks):
(i)~$2~\le~n_{\min}~\le~5$\,---\,\emph{AncSetFit} \cite{pauli-etal-2023-anchoring}, an anchor-based few-shot method that leverages human-readable class descriptions and triplet-loss contrastive learning;
(ii)~$5~<~n_{\min}~\le~80$\,---\,\emph{SetFit} \cite{tunstall2022efficientfewshotlearningprompts}, a sentence-transformer--based few-shot learner with a logistic-regression head;
(iii)~$n_{\min}~>~80$\,---\,full transformer fine-tuning \cite{wolf-etal-2020-transformers} with Optuna-driven \cite{akiba2019optunanextgenerationhyperparameteroptimization} hyperparameter optimisation.
This design means that a practitioner can move from two labelled examples per intent to a production-grade classifier without changing a single line of client code.
A second distinctive feature is the integrated \emph{out-of-distribution (OOD) detection} layer.  Each training method has a companion OOD variant---Marginal Mahalanobis distance for the fine-tuning regime, Maximum Softmax Probability for SetFit, and a logit-based ``outOfScope'' class option---all selectable through a single \texttt{ood\_method} configuration flag.
Third, \name{} ships with an LLM-powered \emph{data augmentation and synthetic test generation} subsystem: when the number of labelled examples is low, the pipeline can call an external language model to synthesise additional training or evaluation samples, guided by automatic domain analysis.

\subsection{Structure}

Figure~\ref{fig:architecture} shows the high-level module layout.  At the top level, the \texttt{auto\_classes} module exposes the four public pipeline classes.  Each pipeline inherits from \texttt{AbstractTrainingPipeline} or \texttt{AbstractInferencePipeline}, which manage the lifecycle stages: data loading, data processing, (optional) data quality evaluation, method resolution, training, evaluation, and model export.

The \texttt{methods} module contains the training algorithms, each subclassing a shared \texttt{Method} base class:
\begin{itemize}
    \item \texttt{Finetuner} / \texttt{FinetunerWithOOD}\,---\,standard transformer fine-tuning with early stopping and optional Optuna HPO.
    \item \texttt{SetFitMethod} / \texttt{SetFitOOD}\,---\,contrastive sentence-transformer training followed by a logistic-regression classifier head.
    \item \texttt{AncSetFitMethod} / \texttt{AncSetFitOOD}\,---\,anchor-label--augmented variant of SetFit for extreme few-shot scenarios.
    \item \texttt{TokenClassificationFinetuner}\,---\,BIO-tagging--based NER with entity-level evaluation.
\end{itemize}
The \texttt{data} module supplies task-specific data providers (\texttt{SimpleDataProvider} for classification, \texttt{SimpleNerDataProvider} for NER) together with an Augmentex-based \cite{martynov2023methodologygenerativespellingcorrection} character- and word-level augmentation engine.  The \texttt{llm\_pipelines} module adds LLM-driven data generation, domain analysis, and synthetic test set construction.

\paragraph{Data quality}
A dedicated \texttt{data\_quality} module implements a pluggable evaluator framework.  A \texttt{DynamicTuner} trains a model on the training set and records per-sample logits across epochs.  Four evaluators consume these signals: \textbf{Retag} \cite{van-halteren-2000-detection}\,--- \, flag samples whose predicted label disagrees with the annotated label, revealing likely annotation errors. \textbf{Uncertainty}\,---\,identifies samples for which the model's softmax probability on the gold class falls below a configurable threshold, indicating ambiguity. \textbf{V-Information}~\cite{ethayarajh2025understandingdatasetdifficultymathcalvusable}\,---\,measures the usable information each sample contributes by comparing the trained model's loss with that of a \emph{null model} (input replaced by a blank token), flagging low-signal examples. \textbf{Dataset Cartography}~\cite{swayamdipta-etal-2020-dataset}\,---\,(text classification only) computes per-sample \emph{confidence} and \emph{variability} across training epochs, partitioning the data into \emph{easy-to-learn}, \emph{ambiguous}, and \emph{hard-to-learn} regions and producing a visual data map. We characterize \emph{hard-to-learn} samples as those highly likely to contain annotation errors. We empirically determined optimal thresholds for the \emph{hard-to-learn} region. 

For token-classification tasks, a \textbf{Label Aggregation} evaluator based on Dawid--Skene~\cite{CrowdKit} consensus estimation is used instead: it runs Monte Carlo dropout to obtain multiple ``annotator'' predictions and aggregates them to detect per-token annotation disagreements.
These evaluators can be run independently via the \texttt{diagnose()} method of any training pipeline, returning a \texttt{DatasetEvaluatorOutput} with per-sample scores and a filtered dataset.

\paragraph{Named Entity Recognition.}
The NER pipeline accepts two annotation formats -- offset-based (start/end character indices) and bracket-based (inline markup) -- and converts both to a BIO tagging scheme internally.  Stratified train/test splitting is performed at the entity level to preserve label proportions.  Evaluation uses \texttt{nervaluate}-based \cite{Batista_nervaluate_2025} entity-level precision, recall, and F1, with support for partial entity matching.

\paragraph{Optimized-inference-ready model serialization}
All methods support ONNX export via \texttt{SaveFormat.ONNX}. The resulting model package bundles the ONNX graph, tokeniser files, label mapping, and a \texttt{meta.json} descriptor.  The inference managers automatically detect available hardware (CUDA, CoreML, CPU) and run batched inference with automatic batch-size detection to prevent out-of-memory failures.

\subsection{Text Classification Training Pipeline}

\paragraph{Data Quality}
Training corpora often contain mislabeled or uninformative examples that degrade model performance\cite{swayamdipta-etal-2020-dataset}. \name{} includes an \textbf{optional} data-quality stage that identifies such samples \emph{before} training using an ensemble of diagnostic methods: \textbf{dataset cartography}~\cite{swayamdipta-etal-2020-dataset}, which tracks per-sample confidence and variability across training epochs and \textbf{v-usable information}~\cite{ethayarajh2025understandingdatasetdifficultymathcalvusable}, which measures how much learnable signal each example carries. We also use \textbf{uncertainty quantification} and \textbf{retagging}, which flag samples with high predictive uncertainty or model-label disagreement.

\paragraph{Objective and search strategy.}
For the fine-tuning regime ($n_{\min}>80$), the pipeline optimises the macro-averaged F1 score on a held-out validation split (90/10 stratified  by default, can be configured), if user explicitly turns hyperparameter optimisation on.  Hyperparameter search is performed with Optuna using a Tree-structured Parzen Estimator (TPE) sampler \cite{watanabe2025treestructuredparzenestimatorunderstanding} over a configurable search space that includes learning rate (log-uniform in $[10^{-6}, 10^{-3}]$), per-device batch size, and weight decay, with a default budget of 10 trials.  Within each trial, early stopping with patience of 5 evaluation steps prevents over-training; the best checkpoint is retained automatically.

For few-shot regimes (SetFit and AncSetFit), no explicit hyperparameter search is conducted: the methods use well-tuned defaults (e.g.\ 20 contrastive iterations, backbone learning rate $10^{-5}$), and training is performed on the full (downsampled) dataset.

\paragraph{Data-level optimisation}
Before method selection, the pipeline performs an adaptive rebalancing step.  If the fraction of low-resource classes (those with $n \le 80$) exceeds a configurable threshold (default 0.3), underrepresented classes are upsampled to $n = 81$ using either Augmentex character/word perturbations or, when enabled, LLM-generated paraphrases. After upsampling, the method resolver is re-evaluated, typically promoting the dataset into the fine-tuning regime.  Conversely, when a few-shot method is selected, overrepresented classes are downsampled to the method's ceiling (80 for SetFit, 5 for AncSetFit) to maintain balanced training.

\paragraph{OOD detection optimisation}
When an OOD-enabled variant is selected, the pipeline jointly optimises the classifier and the OOD detector.  Synthetic out-of-scope samples are generated by a gibberish dataset generator (a function that produces random sequences of randomly placed letters, imitating words in sentences), and a threshold on the chosen OOD score (Mahalanobis distance or maximum softmax probability) is tuned on the validation data.  A user-controllable \texttt{threshold\_factor} (default 1.0) allows for tuning the True-positive rate.

\paragraph{LLM-based test set generation}
In many practical scenarios, users lack a held-out test set to estimate the quality of the model after training. \name{} addresses this by offering an optional LLM-powered test set generation module: given the training data, an LLM synthesizes realistic labeled examples that can serve as a proxy evaluation set. This is particularly valuable in low-resource settings where collecting and annotating additional data is costly. Our experiments (Appendix~\ref{tab:appendix_llm_test_gen}) show that the evaluation scores on the generated test set closely follow those on a real held-out set (with absolute differences below 5 percentage points of up to 80 examples per class), making the generated set a reliable quality signal when no ground-truth test data follow available.

\section{Experiments}

We evaluate on four intent-classification datasets — banking77
\cite{casanueva-etal-2020-efficient}, massive \cite{fitzgerald2022massive1mexamplemultilingualnatural}, hwu64 \cite{liu2019benchmarkingnaturallanguageunderstanding}, and snips \cite{coucke2018snipsvoiceplatformembedded} — spanning binary and multi-class label spaces across low- (5–10 examples per class), medium- (81–100), and full-data regimes. All configurations are averaged over three random seeds with standardized framework presets.

Because \name{} provides automatic OOD detection, we use two evaluation protocols: (1) \textit{OOD-aware}, where OOD samples are included in training as an explicit extra label; and (2) \textit{OOD-in-test}, where OOD samples appear only at test time with no dedicated label in train. We compare against AutoIntent \cite{alekseev2025autointent}, AutoGluon \cite{erickson2020autogluon}, LightAutoML \cite{vakhrushev2021lightautoml}, and H2O \cite{ledell2020h2o}, and report F1-macro, F1-in-scope, and F1-OOD. Data splits and implementation details are provided in Appendix \ref{appendix:sampling}.

To isolate AutoML logic from representation differences, we fix pretrained checkpoints where possible:  bert-base-uncased \footnote{https://huggingface.co/google-bert/bert-base-uncased} is used as the default backbone when supported. For embedding-centric pipelines requiring E5-style representations (notably AutoIntent), we use intfloat/multilingual-e5-large-instruct \footnote{https://huggingface.co/intfloat/multilingual-e5-large}.

This choice ensures consistency across frameworks and isolates the contribution of the \name{} logic itself rather than differences in pretrained representations.

\subsection{OOD Unaware Regime}

\begin{table*}[t]
\centering
\footnotesize
\resizebox{\textwidth}{!}{%
    \begin{tabular}{l l *{10}{r}}
    \toprule
    \textbf{Framework} & \textbf{Preset} & 
    \multicolumn{2}{c}{\textbf{Banking77}} & 
    \multicolumn{2}{c}{\textbf{HWU64}} & 
    \multicolumn{2}{c}{\textbf{MASSIVE}} & 
    \multicolumn{2}{c}{\textbf{SNIPS}} \\ 

    \cmidrule(lr){3-4}\cmidrule(lr){5-6}\cmidrule(lr){7-8}\cmidrule(lr){9-10}
    
    & & \textbf{Macro} & \textbf{OOD} & 
        \textbf{Macro} & \textbf{OOD} & 
        \textbf{Macro} & \textbf{OOD} & 
        \textbf{Macro} & \textbf{OOD} \\
    \midrule
    \name & \textit{supervised} & 0.905 & 0.362 & \textbf{0.893} & 0.276 & 0.875 & 0.471 & \textbf{0.928} & \textbf{0.782} \\
    \name & \textit{unsupervised} & \textbf{0.912} & \textbf{0.433} & 0.890 & \textbf{0.378} & \textbf{0.876} & \textbf{0.515} & 0.921 & 0.761 \\
    \midrule
    AutoIntent & \textit{CL \& unsupervised} & 0.869 & - & 0.829 & - & 0.755 & - & 0.786 & - \\
    AutoIntent & \textit{CL \& supervised} & 0.774 & 0.156 & 0.728 & 0.196 & 0.683 & 0.366 & 0.771 & 0.662 \\
    AutoIntent & \textit{CM \& supervised} & 0.819 & 0.179 & 0.728 & 0.196 & 0.683 & 0.366 & 0.829 & 0.707 \\
    \bottomrule
    \end{tabular}  
}
\caption{F1-Macro and OOD F1-Score for frameworks with automatic OOD detection. In unsupervised regime no OOD train samples are provided, for supervised method OOD samples are selected as described in Section \ref{OOD detection evaluation}. 2 best performing presets for AutoIntent are reported (\textit{CL} for \textit{classic-light} and \textit{CM} for \textit{classic-medium}). Full datasets were used. Best results per dataset are shown in \textbf{bold}.}
\label{table:OOD}
\end{table*}

The OOD-unaware regime mimics a realistic production setting where a deployed model inevitably encounters out-of-distribution input, despite having received no explicit OOD supervision during training. Concretely, OOD samples are present at test time, but no OOD label is provided during training, and the reported F1-Macro is computed exclusively over in-domain classes. This makes the metric a direct measure of how well a framework maintains in-domain classification quality under realistic distributional shift.



\name{} achieves the best or tied performance on three of four datasets, with AutoGluon being the only competitor to outperform it — on Banking77 only, and at a considerably higher computational cost (see Figure~\ref{fig:perfomance_vs_efficiency}). A detailed per-framework breakdown, as well as in-scope F1-Macro results across low-, medium-, and full-data regimes under a clean test condition with no OOD samples, are provided in Appendix~\ref{appendix:id-clf-eval}.

\subsection{OOD detection evaluation} \label{OOD detection evaluation}
We categorize the OOD samples by semantic distance from in-domain data as previously suggested \cite{baran2023classical}:
\begin{enumerate}
    \item \textbf{Close-OOD}: Held-out classes from the same macro-category (scenario) as in-domain classes. Only available for hierarchically-labeled datasets.
    \item \textbf{Mid-OOD}: Held-out classes from different macro-categories within the same dataset.
    \item \textbf{Far-OOD}: Samples from a different dataset with different domain (e.g., Banking77 $\leftrightarrow$ HWU64) \cite{zhan2021out}.
    \item \textbf{Very-Far-OOD}: Synthetically generated gibberish text with no semantic content.
\end{enumerate}

OOD samples are uniformly balanced across all semantic distance categories.

The number of OOD samples in the test set is determined by the 95th percentile of the in-domain class size distribution.

For supervised OOD experiments, OOD samples constitute half of the training set size.

Frameworks with native or explicit support for OOD detection (AutoIntent and \name) are evaluated under supervised OOD settings, where OOD samples were provided during training and testing. Moreover, \name{} was evaluated under an unsupervised OOD regime, where no explicit OOD label was provided during training and OOD samples appear only at test time. We further analyze how OOD detection quality changes when OOD examples are introduced as a labeled class during training. The metrics are shown in the Table~\ref{table:OOD} and a detailed description is provided in the Appendix~\ref{appendix:ood-desc}.

\section{Conclusion}
In this work, we present an easy to use NLU-model producing and serving library. We inspected solutions of our competitors, highlighted their points of growth and offered our library as an all-in-one batteries-included low-code NLU solution. Experimentally, we proved our solution to be well-balanced in terms of the cost and quality of the resulting models. We shown that our Out Of Domain layer is not only unique among other Auto ML libraries in its configurability, but also it is state of the art among solutions that have OOD layers implemented.

\section*{Limitations and Future work}
Our data-driven deterministic optimization strategy will be further analyzed by implementing a metamodel that would decide the best combination of training method, augmentation, and OOD method based on more abstract dataset features such as dataset2vec \cite{jomaa2021dataset2veclearningdatasetmetafeatures}.

\bibliography{custom}

@article{erickson2020autogluon,
  title={Autogluon-tabular: Robust and accurate automl for structured data},
  author={Erickson, Nick and Mueller, Jonas and Shirkov, Alexander and Zhang, Hang and Larroy, Pedro and Li, Mu and Smola, Alexander},
  journal={arXiv preprint arXiv:2003.06505},
  year={2020}
}

@inproceedings{ledell2020h2o,
  title={H2o automl: Scalable automatic machine learning},
  author={LeDell, Erin and Poirier, Sebastien and others},
  booktitle={Proceedings of the AutoML Workshop at ICML},
  volume={2020},
  pages={24},
  year={2020}
}

@inproceedings{alekseev2025autointent,
  title={AutoIntent: AutoML for Text Classification},
  author={Alekseev, Ilya and Solomatin, Roman and Rustamova, Darina and Kuznetsov, Denis},
  booktitle={Proceedings of the 2025 Conference on Empirical Methods in Natural Language Processing: System Demonstrations},
  pages={707--716},
  year={2025}
}

@article{vakhrushev2021lightautoml,
  title={Lightautoml: Automl solution for a large financial services ecosystem},
  author={Vakhrushev, Anton and Ryzhkov, Alexander and Savchenko, Maxim and Simakov, Dmitry and Damdinov, Rinchin and Tuzhilin, Alexander},
  journal={arXiv preprint arXiv:2109.01528},
  year={2021}
}

@inproceedings{casanueva-etal-2020-efficient,
    title = "Efficient Intent Detection with Dual Sentence Encoders",
    author = "Casanueva, I{\~n}igo  and
      Tem{\v{c}}inas, Tadas  and
      Gerz, Daniela  and
      Henderson, Matthew  and
      Vuli{\'c}, Ivan",
    editor = "Wen, Tsung-Hsien  and
      Celikyilmaz, Asli  and
      Yu, Zhou  and
      Papangelis, Alexandros  and
      Eric, Mihail  and
      Kumar, Anuj  and
      Casanueva, I{\~n}igo  and
      Shah, Rushin",
    booktitle = "Proceedings of the 2nd Workshop on Natural Language Processing for Conversational AI",
    month = jul,
    year = "2020",
    address = "Online",
    publisher = "Association for Computational Linguistics",
    url = "https://aclanthology.org/2020.nlp4convai-1.5/",
    doi = "10.18653/v1/2020.nlp4convai-1.5",
    pages = "38--45"
}

@misc{fitzgerald2022massive1mexamplemultilingualnatural,
      title={MASSIVE: A 1M-Example Multilingual Natural Language Understanding Dataset with 51 Typologically-Diverse Languages}, 
      author={Jack FitzGerald and Christopher Hench and Charith Peris and Scott Mackie and Kay Rottmann and Ana Sanchez and Aaron Nash and Liam Urbach and Vishesh Kakarala and Richa Singh and Swetha Ranganath and Laurie Crist and Misha Britan and Wouter Leeuwis and Gokhan Tur and Prem Natarajan},
      year={2022},
      eprint={2204.08582},
      archivePrefix={arXiv},
      primaryClass={cs.CL},
      url={https://arxiv.org/abs/2204.08582}, 
}

@misc{liu2019benchmarkingnaturallanguageunderstanding,
      title={Benchmarking Natural Language Understanding Services for building Conversational Agents}, 
      author={Xingkun Liu and Arash Eshghi and Pawel Swietojanski and Verena Rieser},
      year={2019},
      eprint={1903.05566},
      archivePrefix={arXiv},
      primaryClass={cs.CL},
      url={https://arxiv.org/abs/1903.05566}, 
}

@misc{coucke2018snipsvoiceplatformembedded,
      title={Snips Voice Platform: an embedded Spoken Language Understanding system for private-by-design voice interfaces}, 
      author={Alice Coucke and Alaa Saade and Adrien Ball and Théodore Bluche and Alexandre Caulier and David Leroy and Clément Doumouro and Thibault Gisselbrecht and Francesco Caltagirone and Thibaut Lavril and Maël Primet and Joseph Dureau},
      year={2018},
      eprint={1805.10190},
      archivePrefix={arXiv},
      primaryClass={cs.CL},
      url={https://arxiv.org/abs/1805.10190}, 
}

@inproceedings{swayamdipta-etal-2020-dataset,
    title = "Dataset Cartography: Mapping and Diagnosing Datasets with Training Dynamics",
    author = "Swayamdipta, Swabha  and
      Schwartz, Roy  and
      Lourie, Nicholas  and
      Wang, Yizhong  and
      Hajishirzi, Hannaneh  and
      Smith, Noah A.  and
      Choi, Yejin",
    editor = "Webber, Bonnie  and
      Cohn, Trevor  and
      He, Yulan  and
      Liu, Yang",
    booktitle = "Proceedings of the 2020 Conference on Empirical Methods in Natural Language Processing (EMNLP)",
    month = nov,
    year = "2020",
    address = "Online",
    publisher = "Association for Computational Linguistics",
    url = "https://aclanthology.org/2020.emnlp-main.746/",
    doi = "10.18653/v1/2020.emnlp-main.746",
    pages = "9275--9293"
}

@misc{ethayarajh2025understandingdatasetdifficultymathcalvusable,
      title={Understanding Dataset Difficulty with $\mathcal{V}$-Usable Information}, 
      author={Kawin Ethayarajh and Yejin Choi and Swabha Swayamdipta},
      year={2025},
      eprint={2110.08420},
      archivePrefix={arXiv},
      primaryClass={cs.CL},
      url={https://arxiv.org/abs/2110.08420}, 
}

@inproceedings{pauli-etal-2023-anchoring,
    title = "Anchoring Fine-tuning of Sentence Transformer with Semantic Label Information for Efficient Truly Few-shot Classification",
    author = "Pauli, Amalie  and
      Derczynski, Leon  and
      Assent, Ira",
    editor = "Bouamor, Houda  and
      Pino, Juan  and
      Bali, Kalika",
    booktitle = "Proceedings of the 2023 Conference on Empirical Methods in Natural Language Processing",
    month = dec,
    year = "2023",
    address = "Singapore",
    publisher = "Association for Computational Linguistics",
    url = "https://aclanthology.org/2023.emnlp-main.692/",
    doi = "10.18653/v1/2023.emnlp-main.692",
    pages = "11254--11264"
}

@misc{tunstall2022efficientfewshotlearningprompts,
      title={Efficient Few-Shot Learning Without Prompts}, 
      author={Lewis Tunstall and Nils Reimers and Unso Eun Seo Jo and Luke Bates and Daniel Korat and Moshe Wasserblat and Oren Pereg},
      year={2022},
      eprint={2209.11055},
      archivePrefix={arXiv},
      primaryClass={cs.CL},
      url={https://arxiv.org/abs/2209.11055}, 
}

@misc{akiba2019optunanextgenerationhyperparameteroptimization,
      title={Optuna: A Next-generation Hyperparameter Optimization Framework}, 
      author={Takuya Akiba and Shotaro Sano and Toshihiko Yanase and Takeru Ohta and Masanori Koyama},
      year={2019},
      eprint={1907.10902},
      archivePrefix={arXiv},
      primaryClass={cs.LG},
      url={https://arxiv.org/abs/1907.10902}, 
}

@inproceedings{wolf-etal-2020-transformers,
    title = "Transformers: State-of-the-Art Natural Language Processing",
    author = "Wolf, Thomas  and
      Debut, Lysandre  and
      Sanh, Victor  and
      Chaumond, Julien  and
      Delangue, Clement  and
      Moi, Anthony  and
      Cistac, Pierric  and
      Rault, Tim  and
      Louf, Remi  and
      Funtowicz, Morgan  and
      Davison, Joe  and
      Shleifer, Sam  and
      von Platen, Patrick  and
      Ma, Clara  and
      Jernite, Yacine  and
      Plu, Julien  and
      Xu, Canwen  and
      Le Scao, Teven  and
      Gugger, Sylvain  and
      Drame, Mariama  and
      Lhoest, Quentin  and
      Rush, Alexander",
    editor = "Liu, Qun  and
      Schlangen, David",
    booktitle = "Proceedings of the 2020 Conference on Empirical Methods in Natural Language Processing: System Demonstrations",
    month = oct,
    year = "2020",
    address = "Online",
    publisher = "Association for Computational Linguistics",
    url = "https://aclanthology.org/2020.emnlp-demos.6/",
    doi = "10.18653/v1/2020.emnlp-demos.6",
    pages = "38--45"
}

@misc{jomaa2021dataset2veclearningdatasetmetafeatures,
      title={Dataset2Vec: Learning Dataset Meta-Features}, 
      author={Hadi S. Jomaa and Lars Schmidt-Thieme and Josif Grabocka},
      year={2021},
      eprint={1905.11063},
      archivePrefix={arXiv},
      primaryClass={cs.LG},
      url={https://arxiv.org/abs/1905.11063}, 
}

@misc{martynov2023methodologygenerativespellingcorrection,
      title={A Methodology for Generative Spelling Correction via Natural Spelling Errors Emulation across Multiple Domains and Languages}, 
      author={Nikita Martynov and Mark Baushenko and Anastasia Kozlova and Katerina Kolomeytseva and Aleksandr Abramov and Alena Fenogenova},
      year={2023},
      eprint={2308.09435},
      archivePrefix={arXiv},
      primaryClass={cs.CL},
      url={https://arxiv.org/abs/2308.09435}, 
}

@article{CrowdKit,
  author    = {Ustalov, Dmitry and Pavlichenko, Nikita and Tseitlin, Boris},
  title     = {{Learning from Crowds with Crowd-Kit}},
  year      = {2024},
  journal   = {Journal of Open Source Software},
  volume    = {9},
  number    = {96},
  pages     = {6227},
  publisher = {The Open Journal},
  doi       = {10.21105/joss.06227},
  issn      = {2475-9066},
  eprint    = {2109.08584},
  eprinttype = {arxiv},
  eprintclass = {cs.HC},
  language  = {english},
}

@software{Batista_nervaluate_2025,
author = {Batista, David and Upson, Matthew Antony},
month = jun,
title = {{nervaluate}},
url = {https://github.com/mantisnlp/nervaluate},
version = {1.1.0},
year = {2025}
}

@misc{watanabe2025treestructuredparzenestimatorunderstanding,
      title={Tree-Structured Parzen Estimator: Understanding Its Algorithm Components and Their Roles for Better Empirical Performance}, 
      author={Shuhei Watanabe},
      year={2025},
      eprint={2304.11127},
      archivePrefix={arXiv},
      primaryClass={cs.LG},
      url={https://arxiv.org/abs/2304.11127}, 
}

@inproceedings{zhan2021out,
  title={Out-of-scope intent detection with self-supervision and discriminative training},
  author={Zhan, Li-Ming and Liang, Haowen and Liu, Bo and Fan, Lu and Lam, Albert YS and Wu, Xiao-Ming},
  booktitle={Proceedings of the 59th Annual Meeting of the Association for Computational Linguistics and the 11th International Joint Conference on Natural Language Processing (Volume 1: Long Papers)},
  pages={3521--3532},
  year={2021}
}

@inproceedings{baran2023classical,
  title={Classical out-of-distribution detection methods benchmark in text classification tasks},
  author={Baran, Mateusz and Baran, Joanna and W{\'o}jcik, Mateusz and Zi{\k{e}}ba, Maciej and Gonczarek, Adam},
  booktitle={Proceedings of the 61st Annual Meeting of the Association for Computational Linguistics (Volume 4: Student Research Workshop)},
  pages={119--129},
  year={2023}
}

@misc{vajjala2025textclassificationllmera,
      title={Text Classification in the LLM Era -- Where do we stand?}, 
      author={Sowmya Vajjala and Shwetali Shimangaud},
      year={2025},
      eprint={2502.11830},
      archivePrefix={arXiv},
      primaryClass={cs.CL},
      url={https://arxiv.org/abs/2502.11830}, 
}

@misc{wang2024adaptablereliabletextclassification,
      title={Adaptable and Reliable Text Classification using Large Language Models}, 
      author={Zhiqiang Wang and Yiran Pang and Yanbin Lin and Xingquan Zhu},
      year={2024},
      eprint={2405.10523},
      archivePrefix={arXiv},
      primaryClass={cs.CL},
      url={https://arxiv.org/abs/2405.10523}, 
}

@inproceedings{van-halteren-2000-detection,
    title = "The Detection of Inconsistency in Manually Tagged Text",
    author = "van Halteren, Hans",
    editor = "Abeille, Anne  and
      Brants, Thorsten  and
      Uszkoreit, Hans",
    booktitle = "Proceedings of the {COLING}-2000 Workshop on Linguistically Interpreted Corpora",
    month = aug,
    year = "2000",
    address = "Centre Universitaire, Luxembourg",
    publisher = "International Committee on Computational Linguistics",
    url = "https://aclanthology.org/W00-1907/",
    pages = "48--55"
}


\appendix
\counterwithin{table}{section} 

\renewcommand{\thetable}{\thesection.\arabic{table}} 


\clearpage

\section*{Appendix}

\section{LLM-generated test set evaluation}

The experiments reported in Table~\ref{tab:appendix_llm_test_gen} were conducted using GPT-4o-mini as the generative backend. The module is not tied to this specific model: \name{} supports any OpenAI API-compatible endpoint, including locally hosted models.

\begin{table}[h]
  \centering
  \footnotesize
  \begin{tabular}{lccc}
  \toprule
  \textbf{N-shot} & \textbf{Original} & \textbf{Generated} & \textbf{$|\Delta|$ (pp)} \\
  \midrule
  $[5,\, 10]$   & 0.667 & 0.699 & +0.032 \\
  $[10,\, 20]$  & 0.655 & 0.637 & -0.017 \\
  $[20,\, 40]$  & 0.779 & 0.759 & -0.02 \\
  $[40,\, 80]$  & 0.808 & 0.790 & -0.018 \\
  \midrule
  $[81,\, 100]$ & 0.709 & 0.561 & -0.148 \\
  Full           & 0.759 & 0.690 & -0.689 \\
  \bottomrule
  \end{tabular}
  \caption{Macro-F1 on original vs.\ LLM-generated test sets. The horizontal rule separates regimes where the generated set is a reliable proxy
  ($|\Delta| < 5$\,pp) from those where it is not.}
  \label{tab:appendix_llm_test_gen}
  \end{table}

\section{Threshold selection}
The regime boundaries $n_{\min}$ = 5 and $n_{\min}$ = 80 were determined empirically across a collection of both publicly available English datasets and internal non-English datasets, spanning a variety of classification tasks including intent recognition, sentiment analysis, and general text classification across multiple domains. For each candidate threshold, we measured macro-F1 performance of each training method (AncSetFit, SetFit, and full fine-tuning) across data regimes and identified the per-class sample counts at which transitioning to a more data-intensive method yielded consistent performance gains. The boundaries reported in this work reflect the values at which these gains were found to be stable across the majority of evaluated datasets and domains. We note that while these thresholds may not be universally optimal for every possible dataset, they provide a robust default that eliminates the need for manual method selection in the vast majority of practical settings.

\section{In-domain classification evaluation} \label{appendix:id-clf-eval}

\begin{table}[htbp]
\centering
\footnotesize
\resizebox{\columnwidth}{!}{%
    \begin{tabular}{l l *{10}{r}}
    \toprule
    \textbf{Framework} & 
    {\textbf{Banking77}} & 
    {\textbf{HWU64}} & 
    {\textbf{MASSIVE}} & 
    {\textbf{SNIPS}} \\ 
    
    \midrule
    \name & 0.912 & 0.890 & \textbf{0.876} & \textbf{0.921}  \\
    \midrule
    AutoGluon & \textbf{0.920} & \textbf{0.902} & 0.861 & 0.763 \\
    AutoIntent & 0.869 & 0.829 & 0.755 & 0.786 \\
    LightAutoML & 0.907 & 0.900 & 0.841 & 0.761 \\
    H2O & 0.645 & 0.642 & 0.701 & 0.778  \\
    \bottomrule
    \end{tabular}  
}
\caption{F1-Macro on full datasets under the OOD-unaware regime. OOD samples are present at test time but excluded from the classification report; scores reflect in-domain classification quality only. Best results per dataset are shown in \textbf{bold}.}
\label{table:metrics_subtask2}
\end{table}

Table~\ref{table:metrics_subtask2} shows that \name{} achieves the best or tied performance on three of four datasets — HWU64, MASSIVE, and SNIPS — while remaining highly competitive on Banking77 (0.914 vs. AutoGluon's 0.920). LightAutoML is the closest general-purpose competitor on Banking77 and HWU64, whereas H2O lags substantially across all benchmarks. The only framework to outperform \name{} on any dataset is AutoGluon, which does so only on Banking77 and at a considerably higher computational cost. AutoIntent, despite being purpose-built for intent classification, falls notably short of \name{} on all four datasets, with the largest gap observed on MASSIVE (0.755 vs. 0.880).

Table~\ref{tab:appendix_inscope_automl_performance} further reports in-scope F1-Macro under a stricter controlled condition where no OOD samples appear in the test set, spanning low- (5–10 shots), medium- (40–80 shots), and full-data regimes. These results are broadly consistent with the OOD-unaware findings: \name{} performs strongly in medium- and full-data regimes, particularly on MASSIVE and SNIPS. AutoGluon and LightAutoML are competitive at larger data sizes but degrade significantly in the 5–10 shot setting, while AutoIntent shows the inverse pattern — strongest in low-resource conditions but underperforming as training set size grows. H2O remains the weakest framework across all settings and datasets.

\begin{table*}[t]
\footnotesize
\resizebox{\textwidth}{!}{
    \begin{tabular}{l *{12}{r}}
    \toprule
    \textbf{Framework} & 
    \multicolumn{3}{c}{\textbf{Banking77}} & 
    \multicolumn{3}{c}{\textbf{HWU64}} & 
    \multicolumn{3}{c}{\textbf{MASSIVE}} & 
    \multicolumn{3}{c}{\textbf{SNIPS}} \\ 
    \cmidrule(lr){2-4}\cmidrule(lr){5-7}\cmidrule(lr){8-10}\cmidrule(lr){11-13}
    
    & \textbf{5-10} & \textbf{40-80} & \textbf{Full} & 
      \textbf{5-10} & \textbf{40-80} & \textbf{Full} & 
      \textbf{5-10} & \textbf{40-80} & \textbf{Full} & 
      \textbf{5-10} & \textbf{40-80} & \textbf{Full} \\
    \midrule
    \name{} & 0.727 & 0.881 & 0.920 & 0.747 & 0.753 & 0.902 & \textbf{0.666} & 0.711 & \textbf{0.886} & 0.607 & \textbf{0.890} & \textbf{0.953} \\
    \midrule
    AutoIntent & \textbf{0.808} & 0.848 & 0.782 & \textbf{0.782} & 0.830 & 0.737 & 0.660 & 0.751 & 0.689 & \textbf{0.813} & 0.826 & 0.789 \\
    AutoGluon & 0.598 & \textbf{0.911} & \textbf{0.935} & 0.625 & \textbf{0.899} & \textbf{0.923} & 0.508 & \textbf{0.824} & 0.885 & 0.696 & 0.868 & 0.915 \\
    LightAutoML & 0.524 & 0.891 & 0.922 & 0.555 & 0.891 & 0.920 & 0.425 & \textbf{0.824} & 0.865 & 0.659 & 0.867 & 0.913 \\
    H2O & 0.159 & 0.612 & 0.654 & 0.272 & 0.632 & 0.652 & 0.398 & 0.616 & 0.635 & 0.549 & 0.810 & 0.908 \\
    \bottomrule
    \end{tabular}  
}
\caption{In-scope macro-F1 score comparison across different N-shot settings for intent classification task. Default presets for all frameworks are used.}
\label{tab:appendix_inscope_automl_performance}
\end{table*}

\section{OOD detection description}  \label{appendix:ood-desc}

Table~\ref{table:OOD} reveals several notable patterns across frameworks and regimes. \name{} in the unsupervised regime achieves the best overall balance between in-domain and OOD detection performance, leading on both Macro and OOD F1 on Banking77, HWU64, and MASSIVE. Interestingly, introducing labeled OOD samples during training does not consistently improve \name{}'s performance: while the supervised regime yields the highest OOD F1 on SNIPS (0.782), it comes at the cost of reduced scores on the remaining datasets. This suggests that \name{}'s unsupervised OOD mechanism is already effective without explicit OOD supervision.

AutoIntent requires supervised OOD samples to produce any OOD predictions at all, and even then falls substantially behind \name{} — with OOD F1 gaps as large as 0.304 on Banking77 and 0.144 on MASSIVE. Moreover, AutoIntent's supervised presets incur a significant drop in Macro F1 relative to its unsupervised baseline, indicating that OOD supervision interferes with its in-domain classification. Overall, \name{} provides a more robust and consistent solution for joint in-domain classification and OOD detection across all evaluated datasets and regimes.


\section{Sampling Implementation Details} \label{appendix:sampling}

We evaluate on four English intent classification datasets: Banking77, HWU64, MASSIVE, and SNIPS. For each, we use the same in-distribution versus out-of-distribution (OOD) setup and the same OOD evaluation protocol.

\paragraph{In-distribution sampling.}
We apply a flat in-distribution vs.\ OOD sampling regime. Only classes with at least $n_{\min}$ training examples are retained, where $n_{\min}$ is set to the upper bound of the few-shot range in few-shot runs and to 100 in full-data runs. Among these classes, a fixed fraction (80\%) is chosen at random as in-distribution; the remainder are treated as mid OOD and never appear in training. Within each in-distribution class, 90\% of examples are used for training and 10\% for testing. In few-shot experiments, the training set is further reduced by sampling a random number of examples per class within the specified $n$-shot range; classes that fall below the minimum are dropped.

\paragraph{OOD evaluation setup.}
The test set is augmented with three OOD categories. \emph{Mid OOD} consists of held-out classes from the same dataset (the 20\% of classes not selected as in-distribution). \emph{Far OOD} is drawn from a different intent dataset in the same language to simulate cross-domain OOD; the pairing is chosen so that the far OOD source is semantically related but distributionally distinct (see Table~\ref{tab:far-ood}). \emph{Very far OOD} comprises 1,000 synthetically generated English gibberish utterances to probe robustness to nonsensical inputs. The total number of OOD examples in the combined test set is set to the 95th percentile of in-distribution class sizes; this budget is split equally among the three OOD types. Optionally, OOD examples can also be included in the training set (with total OOD volume equal to half the in-distribution training size, again split across the three types) to assess methods that require OOD at train time.

\begin{table}[t]
\centering
\caption{Far OOD source for each primary intent benchmark.}
\label{tab:far-ood}
\begin{tabular}{ll}
\toprule
Primary dataset & Far OOD source \\
\midrule
Banking77 & HWU64 \\
HWU64 & Banking77 \\
MASSIVE & Banking77 \\
SNIPS & Banking77 \\
\bottomrule
\end{tabular}
\end{table}


\section{Hardware}
All experiments were run on a single machine with the following specifications: an Intel Xeon Gold 6448H processor (64 cores), an NVIDIA H100 GPU with 80 GB VRAM, and 756 GB of system RAM. Training time measurements reported in Figure~\ref{fig:perfomance_vs_efficiency} reflect wall-clock time on this hardware configuration. GPU acceleration was used for frameworks that support it; frameworks without GPU support were run on CPU.

\end{document}